\documentclass[runningheads]{llncs}
\usepackage[T1]{fontenc}
\usepackage{tikz}
\usepackage{multirow}
\usepackage{cite}
\usepackage{comment}
\usepackage{booktabs}
\usepackage{amssymb}
\usepackage{amsmath}
\usepackage{graphicx}
\usepackage{hyperref}
\hypersetup{
    colorlinks=true,
    linkcolor=blue,
    urlcolor=blue,
    citecolor=blue
}
\usepackage[capitalise]{cleveref}
\usepackage{pifont}
\newcommand{\cmark}{\ding{51}}%
\newcommand{\xmark}{\ding{55}}%
\usepackage[x11names]{xcolor}
\usepackage{color}

\begin{document}
\title{DINOLight: Robust Ambient Light Normalization with Self-supervised Visual Prior Integration}
\titlerunning{DINOLight}
%
\author{Youngjin Oh\inst{1}\orcidID{0009-0003-2932-2417} \and
Junhyeong Kwon\inst{1}\orcidID{0009-0003-6383-8775} \and
Nam Ik Cho\inst{1}\orcidID{0000-0001-5297-4649}}
\authorrunning{Oh et al.}
%
\institute{Department of ECE, INMC, Seoul National University, Seoul, Korea\\
\email{\{yjymoh0211,gjh8760,nicho\}@snu.ac.kr}}
\maketitle              
\begin{abstract}
This paper presents a new ambient light normalization framework, DINOLight, that integrates the self-supervised model DINOv2's image understanding capability into the restoration process as a visual prior. Ambient light normalization aims to restore images degraded by non-uniform shadows and lighting caused by multiple light sources and complex scene geometries. We observe that DINOv2 can reliably extract both semantic and geometric information from a degraded image. Based on this observation, we develop a novel framework to utilize DINOv2 features for lighting normalization. First, we propose an adaptive feature fusion module that combines features from different DINOv2 layers using a point-wise softmax mask. Next, the fused features are integrated into our proposed restoration network in both spatial and frequency domains through an auxiliary cross-attention mechanism. Experiments show that DINOLight achieves superior performance on the Ambient6K dataset, and that DINOv2 features are effective for enhancing ambient light normalization. We also apply our method to shadow-removal benchmark datasets, achieving competitive results compared to methods that use mask priors. Codes will be released upon acceptance.

\end{abstract}
\section{Introduction}
\label{sec:intro}
Ambient Light Normalization (ALN)~\cite{vasluianu2024towards} is an emerging challenge in image restoration, extending beyond conventional shadow removal. While most shadow removal techniques focus on eliminating shadows created by a single light source on a flat surface \cite{le2019shadow,wang2018stacked}, ALN addresses a more generalized and challenging scenario where images suffer from multiple shadows resulting from various light sources, complex scene geometries, and diverse surface materials~\cite{vasluianu2024towards}. ALN is particularly relevant to real-world applications, where environmental lighting conditions and scenes are rarely uniform or flat, and shadows can significantly degrade image quality.

\begin{figure}[t]
\centering
   \includegraphics[width=12cm]{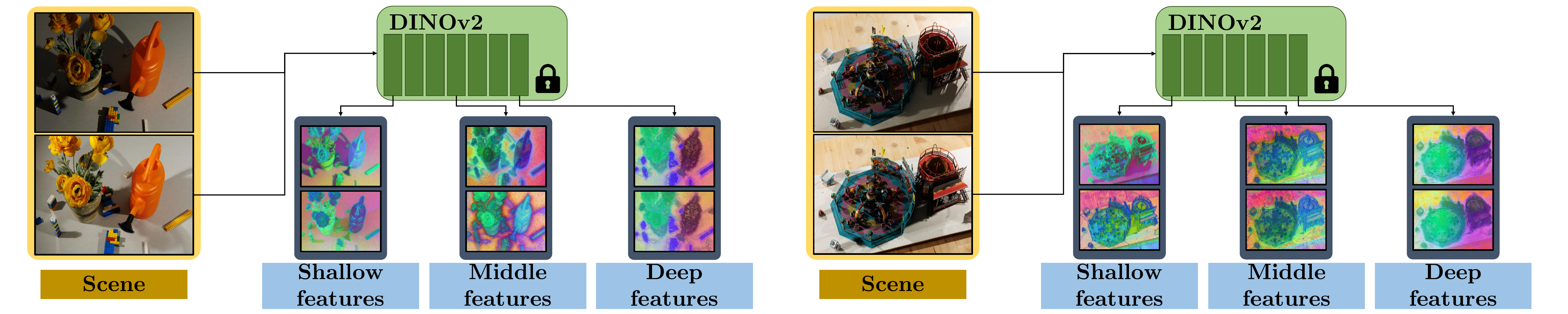}
   \hfil
\caption{Comparison of PCA-computed DINOv2 features of pairs of images from the Ambient6K dataset, which have the same content under different lighting conditions. We observe that DINOv2 features contain degradation-dependent and -independent information that varies with layer depth, motivating us to use them for restoration. More examples can be found in the Supplementary Material.}
\label{figure:figure1}
\end{figure}

A major challenge in shadow removal is identifying the shadowed areas and estimating their extent of degradation. Shadow removal methods, for example,~\cite{le2019shadow,wang2018stacked,cun2020towards,vasluianu2021shadow,vasluianu2023wsrd,fu2021auto,guo2023shadowformer,guo2023shadowdiffusion,vasluianu2024ntire}, address this problem by using shadow mask priors to guide the restoration process. However, these mask priors are not always easy to obtain or accurately estimate in real-world scenarios, which limits the effectiveness of such methods. The reliance on shadow masks can hinder performance, especially in uncontrolled environments where light and shadow are irregular and constantly changing, such as in ALN. This highlights the need for a more reliable and robust approach that does not rely on ground-truth mask information.

In this paper, we note that one potential solution lies in the utilization of visual foundation models. Specifically, we focus on DINOv2~\cite{oquab2023dinov2}, a self-supervised ViT~\cite{dosovitskiy2020image} model, which has demonstrated impressive ability to extract rich and comprehensive features from images, such as geometric information and 3D-aware representations from a single scene~\cite{el2024probing,man2025lexicon3d}, as well as semantic information. This is because the self-supervised nature of DINOv2 allows it to learn from large-scale uncurated datasets, making it a strong choice for tasks that require robust feature extraction.

To utilize DINOv2 as a visual prior for ALN, we first examine its features of degraded images produced by different light sources~\cite{vasluianu2024towards} as shown in \cref{figure:figure1}.
Through observations with many different degraded image pairs, we have found that relatively fine-grained positional and geometric information is generally available in the shallow layers of the ViT-based~\cite{dosovitskiy2020image} DINOv2, while deeper layers tend to generate semantic-aware features~\cite{ranftl2021vision,raghu2021vision,caron2021emerging,oquab2023dinov2}. These characteristics are an ideal prior for ALN, as the extracted features can implicitly provide a deeper understanding of the scene geometry, structure, semantics, and underlying causes of degradation. However, the challenge of effectively integrating the rich information from DINOv2 features into restoration remains non-trivial.

In this context, we propose DINOLight, a new framework that seamlessly embeds the self-supervised visual priors, which are DINOv2 features, into the ALN process. Our approach begins by extracting DINOv2 features from several layers of the ViT DINOv2 model. We then combine these features with our proposed Adaptive Feature Fusion Module (AFFM) (\cref{subsec:affm}). The AFFM is a simple yet effective module that generates point-wise fusion weight masks, which are applied to fuse the DINOv2 features.
The blended features are further integrated with the internal features of our proposed Transformer-based restoration network through a novel Auxiliary Cross-Attention (ACA) mechanism that operates in both the spatial and frequency domains (\cref{subsec:aca}). This dual-domain attention approach, which utilizes DINOv2 features, supplements the self-attention features derived from internal training data. This strategy enhances ALN by incorporating external knowledge learned from large-scale datasets.

Experimental results on the ALN benchmark dataset Ambient6K demonstrate that DINOLight outperforms previous methods in ALN. Additionally, DINOLight achieves favorable results on shadow removal benchmarks when compared to methods that rely on shadow masks. These results suggest that a comprehensive understanding of both geometric and semantic aspects of a scene is beneficial to ALN and shadow removal.

The main contributions of this paper can be summarized as follows:
\begin{itemize}
\item We introduce DINOLight, a novel method that leverages the robust feature-extraction capabilities of DINOv2 for ALN. We believe this is a significant advancement in addressing the complexities of real-world lighting conditions.
\item We present the Adaptive Feature Fusion Module (AFFM), which flexibly merges features extracted from different layers of DINOv2. These fused features are then used as prior information to aid the restoration process through the Auxiliary Cross-Attention (ACA) mechanism.
\item We demonstrate the effectiveness of the proposed method in ALN and show its generalization capability by achieving competitive performance in the closely related task of shadow removal without using shadow masks, suggesting its potential as a unified solution for lighting restoration.
\end{itemize}

\section{Related Work}
\label{sec:related}

\subsection{Shadow Removal}
Shadow removal has been extensively studied in computer vision, with early approaches relying on physical models of shadow formation and handcrafted priors~\cite{arbel2010shadow,liu2008texture,finlayson2002removing,finlayson2009entropy,zhang2015shadow}. Recent advances in deep learning-based image restoration have fostered significant developments in shadow removal as well, with methods ranging from generative models~\cite{le2020shadow,cun2020towards,wang2018stacked,jin2021dc,jin2024des3} and Transformers~\cite{zhang2022spa,guo2023shadowformer,dong2024shadowrefiner,xiao2024homoformer} to learn shadow removal directly from paired shadow and shadow-free images, with only unpaired shadowed images~\cite{hu2019mask,vasluianu2021shadow,liu2021shadow}, or even diffusion~\cite{guo2023shadowdiffusion,mei2024latent,guo2023boundary}. 

Among these, many are mask-prior methods~\cite{le2019shadow,wang2018stacked,vasluianu2021shadow,vasluianu2023wsrd,fu2021auto,guo2023shadowformer,guo2023shadowdiffusion}, which utilize explicit shadow masks to guide the removal process. This leads to more accurate reconstructions, especially in well-defined shadow regions. However, these approaches often struggle with soft shadows and ambiguous mask boundaries, and accurately collecting shadow masks is a demanding and laborious task.

On the other hand, mask-free methods~\cite{jin2021dc,jin2024des3,zhang2022spa,dong2024shadowrefiner} aim to overcome the limitations of mask-prior approaches by removing shadows without explicitly relying on masks. 
Although these mask-free methods moderately remove shadows, they still underperform compared to mask-prior methods, as the masks explicitly guide the networks to focus on areas affected by shadows.

\subsection{Ambient Light Normalization}
A more generalized scenario is ALN~\cite{vasluianu2024towards}, which extends beyond shadow removal to address broader illumination inconsistencies. Unlike traditional shadow removal problems that focus solely on shadows cast by occluders, ALN aims to tackle complex interactions, including self-shadows and varying light sources. IFBlend~\cite{vasluianu2024towards} is a pioneering work in this domain, which defined the task itself. They proposed a robust approach to normalizing lighting conditions across images with diverse lighting variations and a benchmark dataset, Ambient6K, to facilitate research on the topic. By maximizing the joint entropy of image frequencies, IFBlend selectively restores local areas without relying on explicit shadow masks. This scheme is applicable to a broader range of scenarios beyond conventional shadow removal, demonstrating its effectiveness in both structured and unstructured lighting conditions. Concurrent to our work, a challenge~\cite{vasluianu2025ntire,serrano2025promptnorm} was held to foster research on ALN.

\subsection{DINOv2 on Low-Level Vision Tasks}
DINOv2~\cite{oquab2023dinov2} is a self-supervised learning framework designed to produce all-purpose visual features that generalize across diverse image distributions and tasks without fine-tuning. In contrast to CLIP~\cite{radford2021learning}, which is a vision-language model designed to learn visual concepts from natural language supervision that has been frequently utilized for image restoration~\cite{gaintseva2024rave,luo2023controlling,luo2024photo,liang2023iterative,duan2024uniprocessor,lin2024improving}, DINOv2's application to low-level image restoration remains relatively unexplored, albeit its broad adoption for high-level vision tasks.

Recently, promising works on image restoration utilizing DINOv2~\cite{lin2023multi,zhang2025perceive} have been proposed to tackle universal image restoration by exploiting the semantics and robustness of DINOv2 features on multiple degradations. Despite these studies, the application of vision foundation models, such as DINOv2, to other image restoration tasks remains a topic with significant research potential.

While we acknowledge the findings of previous works, we aim to explore additional aspects of DINOv2 for our task. Notably, DINOv2 demonstrates a remarkable ability to capture geometric structures and 3D-aware representations of scenes~\cite{el2024probing,man2025lexicon3d}, as well as to understand semantics. These capabilities are particularly pertinent to ALN, where it is crucial to comprehend the interactions between objects, shadows, textures, and variations in lighting. To our knowledge, this is the first study to utilize DINOv2 features for ALN.

\begin{figure*}[t]
\centering
   \includegraphics[width=12cm]{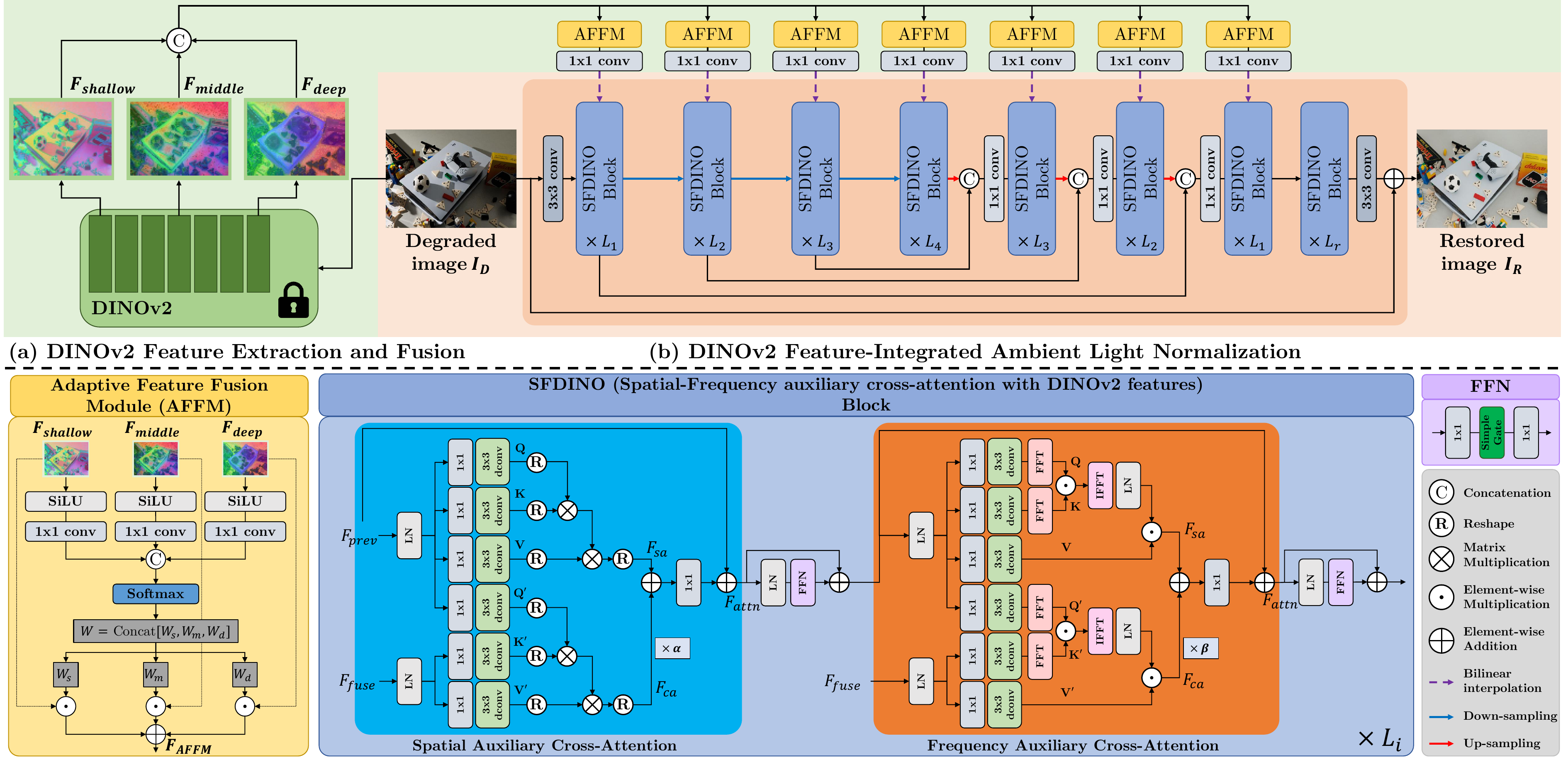}
   \hfil
\caption{Overview of DINOLight, and the two core elements: Adaptive Feature Fusion Module (AFFM), and Auxiliary Cross-Attention (ACA) of the proposed SFDINO block. \textbf{(a)} The first stage involves extracting features from various layers of DINOv2 and combining them using AFFM. \textbf{(b)} In the second stage, the fused features are integrated into the ALN process at multiple resolutions using ACA in SFDINO blocks.}
\label{figure:figure2}
\end{figure*}
\section{Method}
\label{sec:method}

Our primary objective is to develop a novel and effective end-to-end ALN framework that seamlessly integrates self-supervised DINOv2 features as a visual prior across different layers. Specifically, our method is composed of two stages: \textbf{(a)} DINOv2 feature extraction and fusion, and \textbf{(b)} DINOv2 feature-integrated ALN, as illustrated in \cref{figure:figure2}. For each stage, we propose important key components aimed at achieving its specific goal: Adaptive Feature Fusion module (AFFM) for \textbf{(a)}, and Auxiliary Cross Attention (ACA) for \textbf{(b)}.

\subsection{DINOv2 Feature Extraction and Fusion} 
\label{subsec:affm}
The first stage of DINOLight extracts features from different layers of DINOv2 and combines them to prepare for the restoration stage, as shown in \cref{figure:figure2} \textbf{(a)}. We select features from multiple depths of the model as the hierarchical nature of the ViT architecture~\cite{dosovitskiy2020image} allows each layer to progressively capture different types of information.

This phenomenon is evident in the features obtained from degraded images. As illustrated in~\cref{figure:figure1}, shallow features capture positional details and geometric structures, whereas deep features prioritize semantic understanding, focusing on object distinctions rather than fine details retention. Importantly, shallow features reveal lighting-dependent information, while deep features highlight lighting-invariant characteristics. More examples of the features are available in the Supplementary Material.

It is essential to fully utilize these various features to provide optimal information for ALN. In this context, given a degraded image $I_D\in\mathbb{R}^{H\times\ {W}\times {3}}$, we first extract DINOv2 features $F_{shallow},F_{middle},F_{deep}\in\mathbb{R}^{{\frac{H}{14}} \times\ {\frac{W}{14}}\times {768}}$ from layers $1, 6, 12$ of a distilled ViT-B/14 DINOv2 with registers~\cite{dosovitskiy2020image,oquab2023dinov2,darcet2023vision}, and concatenate them. The concatenated features are fused by our proposed AFFM, and the resulting fused features are subsequently passed to the restoration network to enhance its performance.

\noindent{\textbf{Adaptive Feature Fusion Module.}} We develop an efficient and effective strategy that retains the original DINOv2 features while emphasizing the most relevant information at each feature point. First, the concatenated DINOv2 features are used as an input to our AFFM, as illustrated in \cref{figure:figure2}. The features are split into shallow, middle, and deep features ($F_{shallow},\;F_{middle},\;F_{deep}$), and each feature is then processed by a SiLU activation~\cite{elfwing2018sigmoid} followed by a $1\times1$ convolution layer. The outputs are concatenated, and a Softmax function is applied to produce weight maps $W_{s},W_{m},W_{d}\in\mathbb{R}^{{\frac{H}{14}} \times\ {\frac{W}{14}}\times {1}}$. These point-wise fusion weight maps are multiplied to each feature from different layers to emphasize the important components from DINOv2 features containing diverse information that can help locate areas to restore, such as under- or over-lighted areas of a degraded image. The final output $F_{AFFM}$ of AFFM is:
\begin{equation}
    F_{AFFM}=W_{s}\odot F_{shallow}+W_{m}\odot F_{middle}+W_{d}\odot F_{deep},
\label{equation:fusion}
\end{equation}
where $\odot$ denotes element-wise multiplication.

A multi-scale hierarchical structure for an image restoration network~\cite{chen2022simple,zamir2022restormer,wang2022uformer,liang2021swinir} enables processing of input images at different resolutions at each stage, thereby enhancing performance. To exploit the strengths of this design, we use the AFFM to combine DINOv2 features across the hierarchical stages of our restoration network. Each fused feature $F_{AFFM}$ is processed with a $1\times1$ point-wise convolution and bilinear upsampling to ensure consistent feature dimensions across stages. This approach enables us to provide optimal features needed at each stage of the restoration process.

\subsection{DINOv2 Feature-Integrated ALN}
\label{subsec:aca}
In the second stage, DINOLight performs ALN by integrating the fused DINOv2 features generated by AFFM with the restoration network's internal features. To incorporate external features into restoration, we design a Transformer block called ``SFDINO.'' The block is equipped with our proposed attention mechanism, ACA, which effectively utilizes spatial and frequency domains to merge external and internal information. With SFDINO blocks, DINOLight's restoration network is structured to be hierarchical and multi-scale (\cref{figure:figure2} \textbf{(b)}).

\noindent{\textbf{Auxiliary Cross-Attention.}} Self-attention~\cite{vaswani2017attention} operates with key, query, and values generated from self-computed features, and Transformers are used in many restoration tasks~\cite{zamir2022restormer,liang2021swinir} for their excellent performance. We build upon this knowledge and reinforce self-attention with our proposed ACA.

The detailed structure of our SFDINO block is depicted in~\cref{figure:figure2}. For simplicity, we denote the internal features from the previous Transformer block and the fused DINOv2 features for the corresponding stage as $F_{prev}$ and $F_{fuse}\in\mathbb{R}^{\hat{H}\times\ \hat{W}\times \hat{C}}$, respectively.

Our block first applies layer norm on $F_{prev}$ and $F_{fuse}$ to obtain normalized tensors $\textbf{X}$ and $\textbf{X}^\prime$. Following linear transformations $W_Q,W_K,W_V$, and $W_{Q^\prime}$, where each transformation consists of a 1$\times$1 point-wise and a 3$\times$3 depth-wise convolution, we generate query $\textbf{Q}$, key $\textbf{K}$, value $\textbf{V}$, and auxiliary query $\textbf{Q}^\prime$ from $\textbf{X}$. Likewise, we obtain auxiliary key and value $\textbf{K}^\prime$ and $\textbf{V}^\prime$ by applying linear transformations $W_{K^\prime}$ and $W_{V^\prime}$ on $\textbf{X}^\prime$. Next, using transposed attention \cite{zamir2022restormer}, we compute self- and cross-attention using $\textbf{Q}$, $\textbf{K}$, $\textbf{V}$, $\textbf{Q}^\prime$, $\textbf{K}^\prime$, and $\textbf{V}^\prime$. After a channel-wise multiplication with a learnable parameter $\alpha \in\mathbb{R}^{1\times1\times\hat{C}}$ normalized to $[0,1]$ by a Sigmoid function, cross-attention features $F_{ca}$ are added with self-attention features $F_{sa}$ to complement them, hence named as ``Auxiliary Cross-Attention.''
The added features are processed with a 1$\times$1 point-wise convolution and a residual connection. The overall process can be formulated as:
\begin{equation}
    \begin{aligned}
    & \textbf{X}=LayerNorm(F_{prev}),\textbf{X}^\prime=LayerNorm(F_{fuse}), \\
    & \textbf{Q}=W_Q\textbf{X},\textbf{K}=W_K\textbf{X},\textbf{V}=W_V\textbf{X}, \\
    & \textbf{Q}^\prime=W_{Q^\prime}\textbf{X},\textbf{K}^\prime=W_{K^\prime}\textbf{X}^\prime,\textbf{V}^\prime=W_{V^\prime}\textbf{X}^\prime,\\
    & F_{sa} = Attn(\textbf{Q},\textbf{K},\textbf{V}), F_{ca} = Attn(\textbf{Q}^\prime,\textbf{K}^\prime,\textbf{V}^\prime), \\
    & F_{attn} = Conv_{1\times 1}(F_{sa} + \alpha F_{ca})+F_{prev}.
    \end{aligned}
\label{equation:attention}
\end{equation}
The final attention feature $F_{attn}$ is passed to the feed-forward network for further enhancement.
This design allows the ACA to inject complementary context from the fused DINOv2 features into each stage of the network, improving both detail preservation and lighting consistency.

Additionally, inspired by the finding that a frequency-domain approach is effective for ALN~\cite{vasluianu2024towards}, we also compute attentions sequentially using a frequency-domain-based self-attention~\cite{kong2023efficient}. This proposed dual-domain approach complements spatial attention by efficiently capturing long-range dependencies. Network details are provided in the Supplementary Material.

\begin{table*}[t]
\centering
\caption{Quantitative evaluations on Ambient6K test set. We highlight the \colorbox{RosyBrown1}{best}, \colorbox{LightSkyBlue1}{second}, and \colorbox{PaleGreen1}{third best} results. $^{\dagger}$ denotes MACs without additional models.}
\resizebox{0.9\textwidth}{!}{
\begin{tabular}{c|c|c|ccc|c}
\toprule
Method           & Task                                                                                   & Prior     & PSNR$\uparrow$ & SSIM$\uparrow$& LPIPS$\downarrow$ & MACs           \\ \midrule
Degraded         & -                                                                                      & -         & 13.403          & 0.652          & 0.250              & -         \\ \midrule
DCShadowNet~\cite{jin2021dc}      & \multirow{2}{*}{Shadow Removal}                                                        & RGB       & 17.731          & 0.711          & 0.187              & 13.15G    \\
SpA-Former~\cite{zhang2022spa}       &                                                                                        & RGB+Freq. & 19.850          & 0.810          & 0.143              & 16.82G    \\ \midrule
DW-NET~\cite{fu2021dw}           & Dehazing                                                                               & RGB+Freq. & 20.969          & 0.805          & 0.137              & 7.54G     \\ \midrule
NAFNet~\cite{chen2022simple}           & \multirow{7}{*}{Restoration}                                                           & RGB       & 20.580          & 0.808          & 0.142              & 15.92G    \\
MPRNet~\cite{zamir2021multi}           &                                                                                        & RGB       & 20.947          & 0.820          & 0.129              & 37.21G    \\
SFNet~\cite{cui2023selective}            &                                                                                        & RGB+Freq. & 20.519          & 0.812          & 0.141              & 31.27G    \\
SwinIR~\cite{liang2021swinir}           &                                                                                        & RGB       & 20.528          & 0.817          & 0.131              & 37.81G    \\
Uformer~\cite{wang2022uformer}          &                                                                                        & RGB       & 20.776          & 0.818          & 0.131              & 19.33G    \\
Restormer~\cite{zamir2022restormer}        &                                                                                        & RGB       & 21.141          & 0.817          & 0.132              & 35.31G    \\
HINet~\cite{chen2021hinet}            &                                                                                        & RGB       & 20.856          & 0.821          & 0.129              & 42.68G    \\ \midrule
IFBlend~\cite{vasluianu2024towards}          & \multirow{4}{*}{\begin{tabular}[c]{@{}c@{}}Ambient Light\\ Normalization\end{tabular}} & RGB+Freq. & 21.443          & 0.819          & 0.128              & 26.01G    \\
PromptNorm~\cite{serrano2025promptnorm}       &                                                                                        & RGB+Geom. & \colorbox{PaleGreen1}{22.116}          & \colorbox{PaleGreen1}{0.822}          & \colorbox{PaleGreen1}{0.124}              & 13.49G$^{\dagger}$    \\
DINOLight (ours) &                                                                                        & RGB+Freq.+DINO & \colorbox{LightSkyBlue1}{22.788} & \colorbox{LightSkyBlue1}{0.838} & \colorbox{LightSkyBlue1}{0.107}     & 7.89G$^{\dagger}$     \\
DINOLight-L (ours) &                                                                                        & RGB+Freq.+DINO & \colorbox{RosyBrown1}{23.059} & \colorbox{RosyBrown1}{0.839} & \colorbox{RosyBrown1}{0.107}     & 17.48G$^{\dagger}$     \\
\bottomrule
\end{tabular}
\label{table:aln}
}
\end{table*}

\noindent{\textbf{Training Loss.}} 
DINOLight is trained in an end-to-end manner using a combination of $\mathcal{L}_1$ loss and $\mathcal{L}_{MS-SSIM}$ loss~\cite{wang2003multiscale}. Given a ground truth image $I$ and the restored image $I_R$, we minimize the loss defined as:

\begin{equation}
    \mathcal{L}(I,I_R) = |I-I_R|+\lambda(1-\mathcal{L}_{MS-SSIM}(I,I_R)),
\label{equation:loss}
\end{equation}
where $\lambda$ is a balancing hyper-parameter, set as $0.25$.

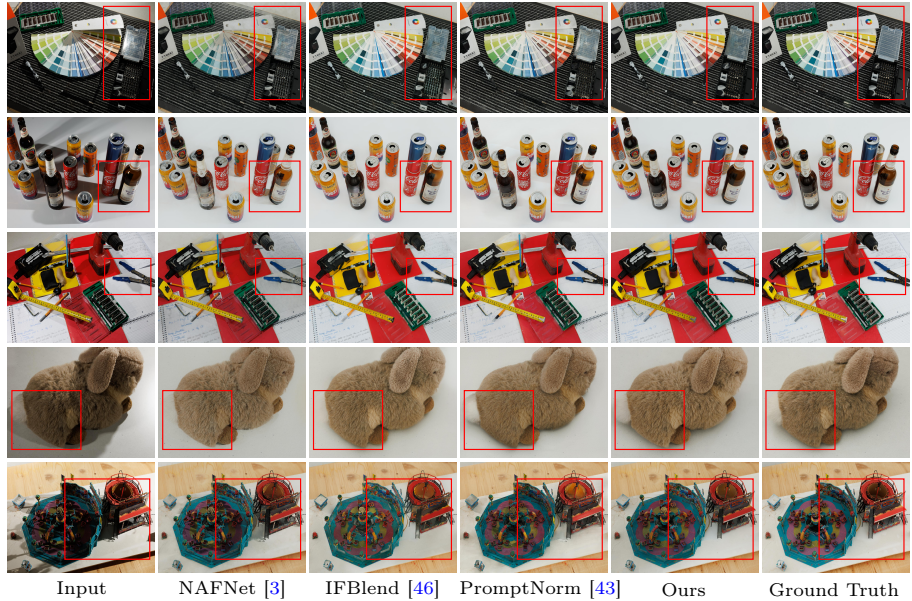
\begin{figure*}[t]
\begin{tikzpicture}{
\scriptsize
    \node(img) at (0,0) {\includegraphics[width=12.0cm]{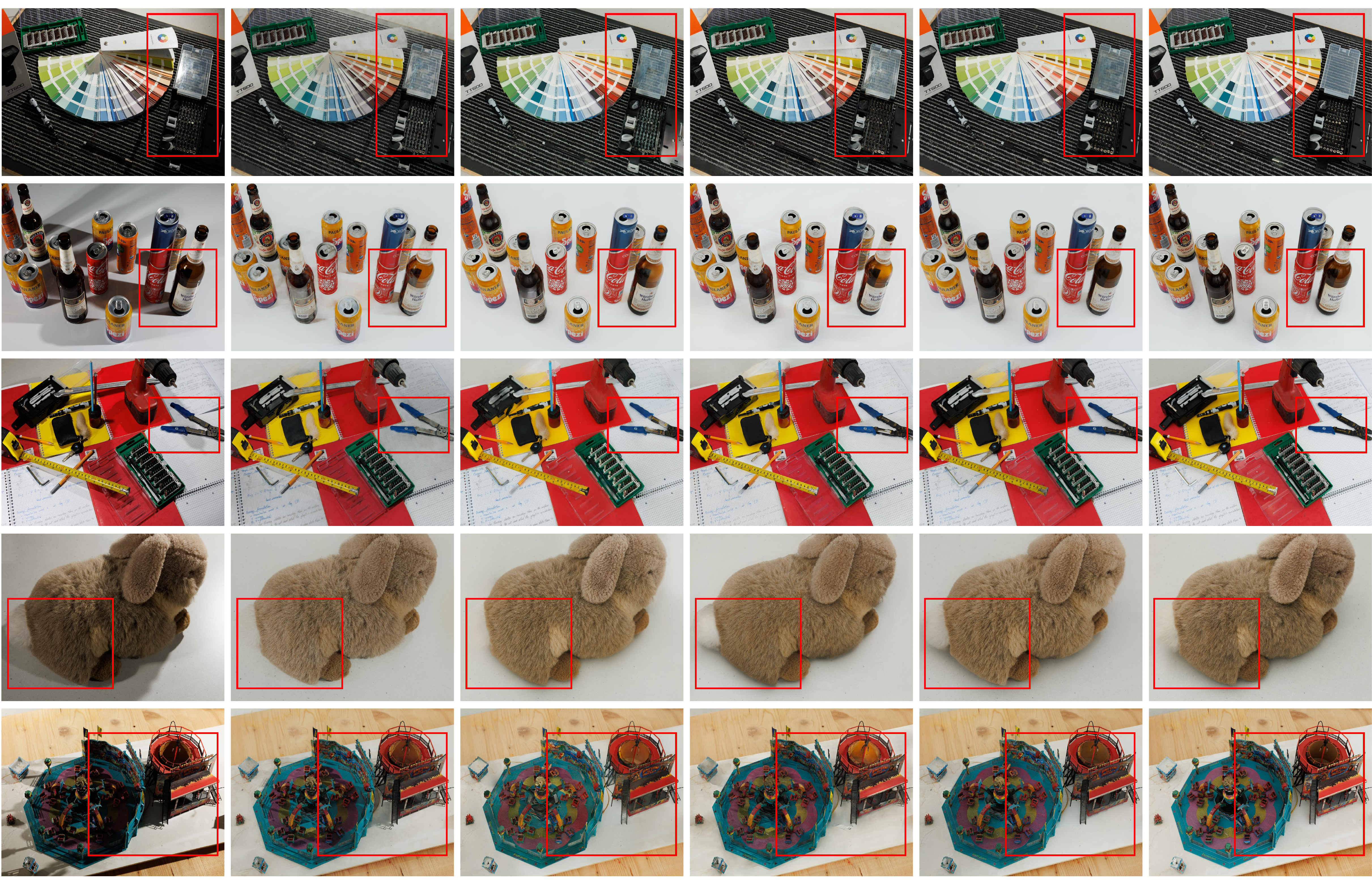}};
    \node(notation_a) at (-5.0,-4.0) {Input};
    \node(notation_c) at (-3.0,-4.0) {NAFNet~\cite{chen2022simple}};    
    \node(notation_d) at (-1.0,-4.0) {IFBlend~\cite{vasluianu2024towards}};    
    \node(notation_e) at (1.1,-4.0) {PromptNorm~\cite{serrano2025promptnorm}};    
    \node(notation_f) at (3.0,-4.0) {Ours};    
    \node(notation_g) at (5.0,-4.0) {Ground Truth};    
}
\end{tikzpicture}

\caption{Qualitative comparison on Ambient6K test set. From left to right: input, restoration results on a versatile restoration network NAFNet~\cite{chen2022simple}, previous ALN methods IFBlend~\cite{vasluianu2024towards} and PromptNorm~\cite{serrano2025promptnorm}, ours, and ground truth. Red boxes highlight regions where DINOLight effectively normalizes light in comparison.}
\label{figure:figure3}
\end{figure*}
\section{Experiments}
\label{sec:experiments}

\noindent{\textbf{Dataset.}} For training and evaluation of DINOLight, we use the ALN benchmark dataset Ambient6K~\cite{vasluianu2024towards}, which is publicly available and contains 5,000 high-resolution image pairs (2560$\times$1920) for training, and 500 test image pairs (1280$\times$960). 
The training data is cropped into 448$\times$448-sized patches. At inference, we adopt a sliding window inference strategy to process the high-resolution input in fixed-size patches identical to the training resolution.

\noindent{\textbf{Training.}} We implement DINOLight using PyTorch with NVIDIA GeForce RTX 3090 GPUs. We use the Adam optimizer~\cite{kingma2014adam} ($\beta_1=0.9,\beta_2=0.999$) and an initial learning rate of $2e-4$ that is gradually decayed to $1e-6$ using cosine annealing~\cite {loshchilov2016sgdr}, with a batch size of 4 for 400K iterations unless specified otherwise. The training patches are augmented with random flipping or rotations. 

\noindent{\textbf{Compared Methods.}} For a fair and reliable comparison, we focused on methods with publicly accessible implementations and reproducible results, in alignment with Vasluianu \textit{et al.}~\cite{vasluianu2024towards}. Therefore, we compare our method with DCShadowNet \cite{jin2021dc}, SpA-Former \cite{zhang2022spa}, DW-NET \cite{fu2021dw}, NAFNet \cite{chen2022simple}, MPRNet \cite{zamir2021multi}, SFNet \cite{cui2023selective}, SwinIR \cite{liang2021swinir}, Uformer \cite{wang2022uformer}, Restormer \cite{zamir2022restormer}, HINet \cite{chen2021hinet}, IFBlend \cite{vasluianu2024towards}, and PromptNorm \cite{serrano2025promptnorm}.

\subsection{Results}
\noindent{\textbf{Quantitative Comparison.}} \cref{table:aln} shows the compared quantitative results on Ambient6K. The measured metrics are PSNR, SSIM, and LPIPS~\cite{zhang2018unreasonable} to assess the restored image quality. Our proposed method achieves superior performance across all metrics compared to previous methods, demonstrating its strength in both fidelity (+0.67~dB in PSNR) and perceptual quality (0.124$\rightarrow$0.107 in LPIPS).

Additionally, our method offers a more favorable trade-off between computational efficiency and restoration quality compared to existing methods. In terms of MACs, the restoration network of DINOLight requires 7.89 GMACs (16.86 GMACs including ViT-B/14 DINOv2), which is efficient than most methods aside from DW-NET.
DINOLight is also lightweight compared to ALN methods IFBlend~\cite{vasluianu2024towards} and PromptNorm~\cite{serrano2025promptnorm} (383.4M and 20M parameters), with 9.5M parameters (95.0M with DINOv2) while delivering superior performance. We also train a larger version of DINOLight, denoted as DINOLight-L, to approximately match the computational resources of previous methods. Notably, it outperforms the prior state-of-the-art PromptNorm by +0.94~dB in PSNR.

\noindent{\textbf{Qualitative Comparison.}} We provide a visual comparison of various methods applied to the Ambient6K dataset in \cref{figure:figure3}. Additional visual results are presented in the Supplementary Material. As illustrated, ALN is not a straightforward task; it requires a method to simultaneously understand the image locally and globally while applying adaptive light normalization to different regions, object relations, and lighting conditions.

Consequently, general image restoration networks, such as NAFNet~\cite{chen2022simple}, and previous ALN approaches \cite{vasluianu2024towards,serrano2025promptnorm} often struggle to distinguish between shadows and objects, particularly in ambiguous regions. This leads to artifacts and color inconsistencies in such areas of an image.
In contrast, our method successfully identifies the regions that require normalization and restores the image accurately, with fewer artifacts and faithful color (first, second, and third rows).

It is also evident that DINOLight fully leverages the rich 3D-aware and semantic priors provided by DINOv2 features. For instance, the fourth row illustrates that DINOLight understands self-shadowed regions that do not require excessive light normalization. 
While PromptNorm~\cite{serrano2025promptnorm} also benefits from incorporating depth information, we observe that it still exhibits some visible artifacts.
Furthermore, our method demonstrates robust lighting normalization results on a challenging sample compared to other methods (fifth row), where intricate geometry induces complicated shadings and lighting effects.

\subsection{Ablation Studies}
We conduct ablation studies on features from different layers of DINOv2 and on the two key components of DINOLight: AFFM and ACA. For the ablation experiments, we train DINOLight on the Ambient6K training set with a batch size of 2 for 200K iterations to streamline the process.

\begin{table}[t]
\centering
\caption{Ablation of DINOv2 features and proposed AFFM on Ambient6K dataset.}
\resizebox{0.66\textwidth}{!}{
\begin{tabular}{c|cccc|ccc}
\toprule
Method    & $F_{shallow}$ & $F_{middle}$ & $F_{deep}$ & AFFM   & PSNR$\uparrow$ & SSIM$\uparrow$ & LPIPS$\downarrow$ \\ \midrule
Baseline  & \xmark        & \xmark       & \xmark     & \xmark & 19.685          & 0.784           & 0.147              \\ \midrule
Model A   & \cmark        & \xmark       & \xmark     & \xmark & 22.207          & 0.822           & 0.119              \\
Model B   & \xmark        & \cmark       & \xmark     & \xmark & 20.605          & 0.802           & 0.138              \\
Model C   & \xmark        & \xmark       & \cmark     & \xmark & 22.060          & 0.823           & 0.117              \\
Model D   & \cmark        & \cmark       & \cmark     & \xmark & 22.414          & 0.829           & 0.115              \\ \midrule
DINOLight & \cmark        & \cmark       & \cmark     & \cmark & 22.600          & 0.830           & 0.112  \\           
\bottomrule
\end{tabular}
\label{table:aln_affm}
}
\end{table}

\begin{table}[t]
\centering
\caption{Ablation of proposed ACA on Ambient6K dataset.}
\resizebox{0.6\textwidth}{!}{
\begin{tabular}{c|ccc|ccc}
\toprule
Method    & \begin{tabular}[c]{@{}c@{}}DINOv2\\ Feature\end{tabular} & \begin{tabular}[c]{@{}c@{}}Spat.\\ Attn\end{tabular} & \begin{tabular}[c]{@{}c@{}}Freq.\\ Attn\end{tabular} & PSNR$\uparrow$  & SSIM$\uparrow$ & LPIPS$\downarrow$\\ \midrule
Baseline  & \xmark                                                         & \cmark                                                      & \cmark                                                        & 19.685 & 0.784 & 0.147 \\ \midrule
Model 1   & \cmark                                                         & \cmark                                                      & \xmark                                                        & 22.289 & 0.828 & 0.114 \\
Model 2   & \cmark                                                         & \xmark                                                      & \cmark                                                        & 22.487 & 0.829 & 0.113 \\ \midrule
DINOLight-p & \cmark                                                         & \cmark                                                      & \cmark                                                        & 22.337 & 0.827 & 0.115 \\ 
DINOLight & \cmark                                                         & \cmark                                                      & \cmark                                                        & 22.600 & 0.830 & 0.112 \\ 
\bottomrule
\end{tabular}
\label{table:aln_aca}
}
\end{table}

\begin{table*}[t]
\centering
\caption{Quantitative evaluations on ISTD~\cite{wang2018stacked} and ISTD+~\cite{le2019shadow} test dataset. RMSE in the LAB color space is measured as the metric.
Best results are highlighted in bold for each mask-prior and mask-free method, and unavailable results are omitted.}
\resizebox{0.95\textwidth}{!}{
\begin{tabular}{c|c|c|ccc|ccc}
\toprule
\multirow{2}{*}{Method} & \multirow{2}{*}{\begin{tabular}[c]{@{}c@{}}Mask\\ Usage\end{tabular}} & \multirow{2}{*}{\begin{tabular}[c]{@{}c@{}}Evaluation\\ Resolution\end{tabular}} & \multicolumn{3}{c|}{ISTD}                     & \multicolumn{3}{c}{ISTD+}                     \\ \cmidrule{4-9} 
                        &                                                                       &                                                                                  & Total $\downarrow$        & Shadow $\downarrow$       & Shadow-Free $\downarrow$   & Total $\downarrow$         & Shadow $\downarrow$       & Shadow-Free $\downarrow$  \\ \midrule
Degraded                & -                                                                     & 640$\times$480                                                                   & 6.80          & 15.07         & 3.86          & 7.15          & 17.53         & 1.82          \\ \midrule
ST-CGAN~\cite{wang2018stacked}                 & \multirow{7}{*}{Mask-Prior}                                           & 256$\times$256                                                                   & 4.05          & 4.83          & 3.44          & -             & -             & -             \\
SP-M Net~\cite{le2019shadow}                &                                                                       & 512$\times$512                                                                   & -             & -             & -             & 4.37          & 4.79          & 4.27          \\
DHAN~\cite{cun2020towards}                    &                                                                       & 640$\times$480                                                                   & 3.43          & 4.65          & 3.13          & 3.19          & 4.04          & 2.97          \\
PULSr~\cite{vasluianu2021shadow}                   &                                                                       & 512$\times$512                                                                   & 3.33          & 4.48          & 3.03          & 2.82          & 4.12          & 2.39          \\
DNSR~\cite{vasluianu2023wsrd}                    &                                                                       & 640$\times$480                                                                   & 2.84          & 4.39          & 2.47          & 2.24          & 3.92          & 1.80          \\
AEF~\cite{fu2021auto}                     &                                                                       & 256$\times$256                                                                   & 3.10          & 3.75          & 2.79          & 2.31          & 3.23          & 2.05          \\
ShadowFormer~\cite{guo2023shadowformer}            &                                                                       & 640$\times$480                                                                   & \textbf{2.43} & \textbf{3.25} & \textbf{2.38} & \textbf{1.93} & \textbf{2.93} & \textbf{1.66} \\ \midrule
DCShadowNet~\cite{jin2021dc}             & \multirow{7}{*}{Mask-Free}                                            & 640$\times$480                                                                   & 4.06          & 5.97          & 3.62          & 4.60          & 10.30         & 3.50          \\
DeS3~\cite{jin2024des3}                    &                                                                       & 640$\times$480                                                                   & -             & -             & -             & 2.80          & 4.90          & 2.70          \\
SwinIR~\cite{liang2021swinir}                  &                                                                       & 640$\times$480                                                                   & 3.70          & 4.84          & 3.44          & 3.00          & 4.54          & 2.44          \\
Uformer~\cite{wang2022uformer}                 &                                                                       & 640$\times$480                                                                   & 3.72          & 4.83          & 3.47          & 3.17          & 4.50          & 2.71          \\
Restormer~\cite{zamir2022restormer}               &                                                                       & 512$\times$512                                                                   & 3.73          & 5.30          & 3.44          & 3.13          & 4.17          & 2.79          \\
IFBlend~\cite{vasluianu2024towards}                 &                                                                       & 640$\times$480                                                                   & 3.52          & 4.02          & 3.38          & 2.40          & \textbf{3.18}          & 2.14          \\
DINOLight (ours)        &                                                                       & 640$\times$480                                                                   & \textbf{3.00} & \textbf{3.63} & \textbf{2.84} & \textbf{2.34}             & 3.41             & \textbf{1.99}             \\ 
\bottomrule
\end{tabular}
\label{table:sr}
}
\end{table*}

\noindent{\textbf{Impact of DINOv2 features and AFFM.}} \cref{table:aln_affm} shows the ablation results of each shallow, middle, and deep feature of DINOv2, and AFFM. The Baseline refers to a modified DINOLight that does not incorporate DINOv2 features; instead, it uses the internal features $F_{prev}$ to compute cross-attention. Models A, B, and C are models in which each shallow, middle, and deep features from DINOv2 were used independently without fusion with AFFM to compute ACA. Model D is a variant that combines the features using a simple addition.

We observe that all features improve the performance compared to the Baseline, with the greatest gain from $F_{shallow}$, which contains geometrical and positional cues closely related to the degradations caused by multiple light sources. However, using only $F_{shallow}$ is suboptimal compared to Model D and DINOLight, both of which use all three features. This suggests that ALN benefits from the complementary knowledge of degradation-dependent and -independent information across the layers of DINOv2, which is consistent with our underlying motivation.

The effectiveness of our proposed AFFM is further demonstrated by comparing Model D with DINOLight. The results indicate that AFFM contributes to ALN by adaptively selecting and emphasizing relevant features according to the hierarchical stage of the restoration network.

\noindent{\textbf{Effect of ACA.}} \cref{table:aln_aca} highlights the impact of ACA when utilizing DINOv2 features. The Baseline is the same version of DINOLight in \cref{table:aln_affm}. Model 1 relies solely on spatial attention, while Model 2 utilizes only frequency attention. In contrast, DINOLight combines both types of attention in an alternating order, sequentially using SFDINO blocks. Furthermore, we train a variant of DINOLight called DINOLight-p, in which the dual-domain ACA is run in parallel to compare with the sequential setting.

Leveraging DINOv2 features with our proposed module significantly enhances ALN, with a minimum increase of 2.80~dB in PSNR. Additionally, our SFDINO block further enhances restoration fidelity, demonstrating the synergistic effect of the dual-domain approach (DINOLight), compared to a purely spatial (Model 1) or a purely frequency-based approach (Model 2). We provide the visual ablation results in the Supplementary Material for extensive analysis.

In the parallel setting of ACA, DINOLight exhibits inferior performance relative to its deeper sequential counterpart, supporting our design choice. We also report that the ordering of spatial and frequency attentions in a sequential configuration has a negligible influence on performance.

\subsection{Application to Shadow Removal}
To verify the strength of our method, we also train and evaluate DINOLight on shadow removal datasets ISTD~\cite{wang2018stacked} and ISTD+~\cite{le2019shadow}. Since shadow removal can be considered as a specific instance of ALN, DINOLight should also perform robustly for shadow removal. Following the approach in Vasluianu \textit{et al.}~\cite{vasluianu2024towards}, we report the RMSE in the LAB color space for ISTD/ISTD+ datasets, and compared methods for fair comparison. Provided binary shadow masks are applied to separately measure RMSE in the shadow and shadow-free regions, including the total RMSE.

\noindent{\textbf{Experimental Results.}} \cref{table:sr} shows that DINOLight performs competitively to most methods that rely on mask priors, which have the advantage of narrowing down the shadow-affected regions using the shadow masks. Compared to mask-free methods, DINOLight outperforms by a large margin on ISTD and performs favorably on ISTD+, demonstrating that it can generalize to shadow removal as well.
Qualitative results for ISTD+ and additionally WSRD+~\cite{vasluianu2023wsrd} are provided in the Supplementary Material.
\section{Conclusion}
\label{sec:conclusion}
We have developed a novel solution for ALN based on the properties of DINOv2 features. Our proposed framework is two-stage: DINOv2 feature extraction and fusion, followed by ALN using the fused features. At each stage, we designed the AFFM to emphasize relevant features and the ACA mechanism to utilize both the spatial and frequency domains. Extensive experiments on benchmark ALN and shadow-removal datasets demonstrate the effectiveness and generalization of our method in restoration tasks that require a more comprehensive understanding of scene structure, semantics, and lighting conditions.


\bibliographystyle{splncs04}

\end{document}